\documentclass[]{spie}  

\usepackage{amsmath,amsfonts,amssymb}
\usepackage{graphicx}
\usepackage[colorlinks=true, allcolors=blue]{hyperref}
\usepackage{float}
\usepackage{subcaption}
\usepackage{hyperref}
\usepackage{booktabs}

\usepackage{setspace}
\DeclareMathOperator*{\argmax}{arg\,max}

\title{A deep Q-learning method for optimizing visual search strategies in backgrounds of dynamic noise}

\author[a]{Weimin Zhou}
\author[a]{Miguel P. Eckstein}
\affil[a]{Department of Psychological \& Brain Sciences, University of California, Santa Barbara, CA~93106, USA}

\authorinfo{Further author information: (Send correspondence to Weimin Zhou and Miguel P. Eckstein)\\Weimin Zhou: E-mail: weiminzhou@ucsb.edu\\Miguel P. Eckstein: E-mail: miguel.eckstein@psych.ucsb.edu}

\begin{document} 
\maketitle

\begin{abstract}
Humans process visual information with varying resolution (foveated visual system) and explore images by orienting through eye movements the high-resolution fovea to points of interest. The Bayesian ideal searcher (IS) that employs complete knowledge of task-relevant information optimizes eye movement strategy and achieves the optimal search performance. The IS can be employed as an important tool to evaluate the optimality of human eye movements, and potentially provide guidance to improve human observer visual search strategies. Najemnik and Geisler (2005) derived an IS for backgrounds of spatial 1/f noise. The corresponding template responses follow Gaussian distributions and the optimal search strategy can be analytically determined. However, the computation of the IS can be intractable when considering more realistic and complex backgrounds such as medical images. Modern reinforcement learning methods, successfully applied to obtain optimal policy for a variety of tasks, do not require complete knowledge of the background generating functions and can be potentially applied to anatomical backgrounds. An important first step is to validate the optimality of the reinforcement learning method.  In this study, we investigate the ability of a reinforcement learning method that employs Q-network to approximate the IS. We demonstrate that the search strategy corresponding to the Q-network is consistent with the IS search strategy. The findings show the potential of the reinforcement learning with Q-network approach to estimate optimal eye movement planning with real anatomical backgrounds.
\end{abstract}


\section{INTRODUCTION}
\label{sec:intro}  
There is a long tradition of studying eye movements with radiologists \cite{kundel1989searching, kundel2004reader}. Much of that research has focused on measuring eye movements to determine whether radiologist errors are being fixated and missed (recognition errors) or are simply not fixated (search errors).  Other studies have concentrated on assessing how eye movements vary across radiology expertise \cite{manning2006radiologists, bertram2013effect}.

Little is known about what types of eye movement patterns maximize perceptual performance.  For 3D CT images, studies show a relationship between expertise, perceptual performance and a strategy to fixate at a few points and scrolling across slices (drilling strategy) \cite{drew2013scanners}. However, a study \cite{eckstein2018evaluation} has found an interaction between the target type and the effect of the eye movement strategy.  For example, for small targets not visible in the visual periphery, a drilling  can lead to suboptimal performance. 
However, no method has been developed to calculate the performance maximizing eye movements for anatomical backgrounds in medical images or realistic simulations. 

Researchers in vision science have developed a method to calculate optimal eye movement strategies for computer generated noise textures. The ideal searcher (IS) employs complete task-relevant information, background statistics, and makes the next optimal eye movement that maximizes the search performance. Najemnik and Geisler have developed an IS model for a foveated visual system considering backgrounds of $1/f$ noise \cite{najemnik2005optimal}, and studies have demonstrated that the IS can provide important insight into human perceptual processing for search tasks \cite{najemnik2008eye, eckstein2015optimal, hoppe2019multi}.
However, the application of the IS model has been limited to filtered noise backgrounds \cite{najemnik2008eye, ackermann2013choice,paulun2015visual,zhang2010evolution}, from which the optimal fixation selection can be analytically determined based on Gaussian distribution. When more realistic and complicated backgrounds are considered that cannot be described analytically by a simple distribution such as Gaussian, the computation of the IS can be difficult or even intractable.

Machine learning-based methods have recently been actively explored to establish model observers for a variety of tasks such as binary signal detection tasks \cite{zhou2018learning,zhou2019learning_HO,zhou2019approximating, zhou2020markov, he2020learning, kim2020convolutional, jonnalagadda2020evaluation}, joint signal detection-location tasks \cite{zhou2019learningIO, zhou2020approximating}, and joint signal detection-estimation tasks\cite{li2021supervised, li2021hybrid}.
However, it remains unclear how these methods can be applied for performing visual search tasks that can be essentially described by control problems.
Reinforcement learning methods have been actively explored and applied to solve control problems that require an agent to take optimal action \cite{sutton2018reinforcement, mnih2013playing}. Unlike supervised learning-based methods that train models based on labeled training dataset, reinforcement learning methods train models by interacting an agent with an environment. Specifically, the model is trained to take an action such that the reward given by the environment is maximized. The reward represents a quantity that describes how good the action is for the considered task. Recent studies have developed reinforcement learning methods for visual tasks. For example, Mnih et. al developed a recurrent attention model (RAM) that employs REINFORCE algorithm for image classification tasks \cite{mnih2014recurrent} and Florin et. al developed a multi-scale deep Q-learning method for 3D-landmark detection in CT scans \cite{ghesu2017multi}. However, it remains unclear how these methods can be employed to approximate the IS for visual search tasks.

In this study, we propose and investigate a deep Q-learning method to approximate the IS for an model with a foveated visual system performing a visual search task. To validate the deep Q-learning method, we conduct  simulation studies with Gaussian-distributed template responses for which the IS can calculated and compared to the proposed deep Q-learning method. We demonstrate that the eye-movement endpoint (fixations) distribution and the search performance corresponding to the proposed Q-learning method are consistent with those of the IS developed by Najemnik and Geisler.

\section{Background}
\subsection{Ideal searcher in a foveated visual system}
\label{sec:title}

In this study, we consider the IS for a foveated visual system in dynamic backgrounds, i.e., temporally uncorrelated external noise. A foveated visual system  perceives a degraded image representation that are affected by reduced spatial resolution in the visual periphery and neural noise. The performance of the IS model away from its point of fixation is  determined by the target visibility map $d'(\epsilon)$ that is defined as the signal-to-noise ratio of the template responses at different retinal eccentricities $\epsilon$ (distance of the target from the fovea). 

Consider a search task that requires an observer to localize a target that has $n$ possible locations. Let $W_{i, k(t)}$ be the template response at the $i^{th}$ target location from the fixation $t$ having the fixation location $k(t)$. According to the visibility map, $W_{i, k(t)}$ can be described by a distribution having the mean $\mu_{i, k(t)} = 0.5$ if the target is at the location $i$ and, $\mu_{i, k(t)} = -0.5$, otherwise, and the standard deviation $\sigma_{i, k(t)} = \frac{1}{d'_{i, k(t)}}$, where $d'_{i, k(t)}$ denotes the visibility of the target at location $i$ when the fixation $t$ has the location $k(t)$.

Let $P(i)$ denote the prior probability of the target at location $i$, when $W_{i, k(t)}$ can be described a Gaussian distribution, the posterior probability that the target is at the location $i$ after $T$ fixations can be computed as \cite{najemnik2005optimal}:
\begin{equation}
    p_i(T) = \frac{P(i)\exp [\sum_{t=1}^T {d'_{i, k(t)}}^2W_{i, k(t)}]}{\sum_{j=1}^n P(j)\exp [\sum_{t=1}^T {d'_{j, k(t)}}^2W_{j, k(t)}]}
\end{equation}
A maximum a posterior (MAP) searcher \cite{najemnik2005optimal, eckstein2001quantifying} selects the next fixation location $k_{MAP}(t+1)$ as:
\begin{equation}
    k_{MAP}(T+1) = \argmax_{i} p_i(T)
\end{equation}
The ideal searcher selects the next fixation location that will maximize the probability of correctly localizing the target after that fixation \cite{najemnik2005optimal}:
\begin{equation}
    k_{IS}(T+1) = \argmax_{k(T+1)}(\sum_{i=1}^n p_i(T) p(correct | i, k(T+1))),
\end{equation}
where $p(correct | i, k(T+1))$ is the probability of the target at location $i$ being correctly identified, which can be computed as:
\begin{equation}
    p(correct | i, k(T+1)) = p\Big(p_i(T+1)\geq p_1(T+1),...,p_i(T+1)\geq p_n(T+1)|i, k(T+1)\Big).
\end{equation}
When the template response $W_{i, k(t)}$ follows Gaussian distribution, the probability $p(correct | i, k(T+1))$ can be analytically determined as \cite{najemnik2005optimal}:
\begin{equation}
\begin{split}
    p(correct | i, k(T+1)) = \int_{-\infty}^{\infty} d'_{i, k(T+1)} &\phi \left(d'_{i, k(T+1)} (W_{j, k(T+1)} - 0.5)\right) \times \\
    &\prod_{i\neq j}  \Phi \left[\frac{\ln[\frac{p_j(T)}{p_i(T)}]\frac{1}{{d'_{j, k(T+1)}}^2} + \frac{ {d'_{i, k(T+1)}}^2}{ {d'_{j, k(T+1)}}^2} W_{i, k(T+1)}+0.5}{1/d'_{j, k(T+1)}}\right] dW_{i, k(T+1)},
    \end{split}
\end{equation}
where $\phi(\cdot)$ is the standard normal density function and $\Phi(\cdot)$ is the standard normal integral function. 

\subsection{Reinforcement learning and Q-network}
Reinforcement learning (RL) involves training an intelligent agent that interacts with an environment to take actions that maximize a reward. A reward is a numeric value that represents how good the task would be performed by taking an action. The RL problem can be described mathematically by a Markov Decision Process (MDP) \cite{van2012reinforcement, sutton2018reinforcement, hoppe2019multi}, which is defined by a tuple $(S, A, R, P, \gamma)$. Here, $S$ denotes a set of possible states, $A$ denotes a set of possible actions, $R$ denotes the distribution of reward given a pair of state and action,
$P$ denotes the transition probability of the next state given a pair of state and action, and $\gamma$ denotes the discount factor.
The initial state $s_0\in S$ is sampled by the environment at time step $t=0$. Then for $t>0$, the agent selects an action $a_t$ that follows a policy $\pi(a_t|s_t)$, and the environment samples a reward $r_t$ from the distribution $R(\cdot|s_t, a_t)$ and the next state $s_{t+1}$ from the distribution $P(\cdot|s_t, a_t)$ that are returned to the agent. The goal is to train an agent that takes actions following the optimal policy $\pi^*(a_t|s_t)$ to maximize a cumulative discounted reward $R_c =\sum_{t\geq 0}\gamma^t r_t \equiv r_0 + \sum_{t\geq 1}\gamma^t r_t$. 

The Q-value function is defined as the expected cumulative reward if the agent takes action $a$ in state $s$ and then follows the policy $\pi$ \cite{van2012reinforcement, sutton2018reinforcement}:
\begin{equation}
    Q^{\pi}(s,a) = \mathbb{E}\left[\sum_{t\geq 0}\gamma^t r_t | s_0=s,a_0=a, policy=\pi \right].
\end{equation}
The optimal Q-value function $Q^*(s,a)$ is the maximum $Q^{\pi}(s,a)$ over all possible policies:
\begin{equation}
    Q^*(s,a)=\max_{\pi} Q^{\pi}(s,a).
\end{equation}
According to Bellman equation, the $Q^*(s,a)$ can be reformatted as \cite{van2012reinforcement, sutton2018reinforcement}:
\begin{equation}
    Q^*(s,a) = \mathbb{E}_{s'}[r+\gamma \max_{a'}Q^*(s',a')|s,a],
\end{equation}
where $Q^*(s',a')$ is the optimal Q-value function if the agent takes action $a'$ in state $s'$ at the next time-step.
The basic idea of Q-learning is to update the estimated the $Q^*(s,a)$ value by use of Bellman equation \cite{van2012reinforcement, sutton2018reinforcement}:
\begin{equation}
    Q_{i+1}(s,a) = \mathbb{E}[r+\gamma \max_{a'}Q_i(s',a')].
\end{equation}
It can be shown that $Q_i\rightarrow Q^*$ as $i\rightarrow \infty$ \cite{}.
However, this method alone cannot scale well because one must compute the Q-value function for every state-action pairs. 
This problem can be solved by employing a function approximator to estimate the action-value function:
\begin{equation}
    Q(s,a;\theta)\approx Q^*(s,a),
\end{equation}
where $Q(s,a;\theta)$ is a function approximator with weights $\theta$. When a neural network is employed to represent the function approximator, it is referred to as Q-network. Let $y_i$ denote the target Q-value for iteration $i$, $y_i = \mathbb{E}[r+\gamma \max_{a'}Q(s',a'; \theta_{i-1}|s,a)]$, the Q-network can be trained by minimizing the loss function $L_i(\theta_i)$:
\begin{equation}
\label{eq:L}
    L_i(\theta_i) = \mathbb{E}_{s,a\sim \rho(\cdot)}[(y_i - Q(s,a,;\theta_i))^2],
\end{equation}
where $\rho(\cdot)$ is a probability distribution of the state-action pair that is used to explore the state space adequately. This distribution is referred to as the behaviour distribution. 
Once the Q-network is trained, the agent takes an action by use of a greedy strategy: $a = \argmax_{a}Q(s,a;\theta)$.

\section{Deep Q-learning-enabled ideal searcher} 
We consider the ideal searcher that looks one fixation into the future in dynamic external noise. To train a Q-network for learning the ideal searcher, a Markov Decision Process is established as follows:
\begin{itemize}
  \item State: The state is defined in a way that the task-specific information extracted from the past and current observations can be summarized. Specifically, the state after $T$ fixations is defined as a vector $\mathbf{s}_T$ having element $s_T(i)$ as:
  \begin{equation}
      s_T(i) = \sum_{t=1}^T {d'_{i, k(t)}}^2W_{i, k(t)}.
  \end{equation}
  
  \item Actions: There are two actions to be determined. The first is the behaviour action used for estimating the Q-value function. The second is the localization action used for localizing the target. To estimate the Q-value function, for a given state, we evaluate the Q-value for all possible actions corresponding to all possible target locations for a given state. To localize the target, the agent first selects the next fixation $k(T+1)$ from the $n$ possible target locations that maximizes the estimated Q-value: $k(T+1) = \argmax_{a}Q(s,a;\theta)$. Subsequently, template responses from fixation $T+1$ are computed and the state at the next-step $\mathbf{s}_{T+1}$ is evaluated. The target location is determined as the one that has the maximum state: $i^* = \argmax_{i} s_{T+1}(i)$.  
  
  \item Reward: The immediate reward $r$ is set to 1 if $i^*$ corresponds to the ground-truth target location. Otherwise, $r = 0$.  
  Because only one future fixation is considered, the discount factor $\gamma$ is set to 0 and the Q-value function becomes:
  \begin{equation}
  \label{eq:Q}
      Q^*(s,a) = \mathbb{E}_{s'}[r|s,a] = \int r(s') p(s'|s,a) ds'.
  \end{equation}
  The $Q^*(s,a)$ can be estimated by use of Monte Carlo integration: $Q^*(s,a)\approx \sum_{j=1}^{J} r(s'_j)$, where $s'_j$ is sampled from the distribution $p(s'|s,a)$. It should be noted that the environment returns the reward $r(s')$ and the distribution $p(s'|s,a)$ based on the knowledge of ground-truth target location. 
  
  A Q-network can be subsequently trained according to Eq. (\ref{eq:L}) and the next fixation is selected such that the Q-value represented by the Q-network is maximized. 
\end{itemize}

\section{Numerical studies and results}
\begin{figure}[H]
         \centering
         \includegraphics[width=0.7\textwidth]{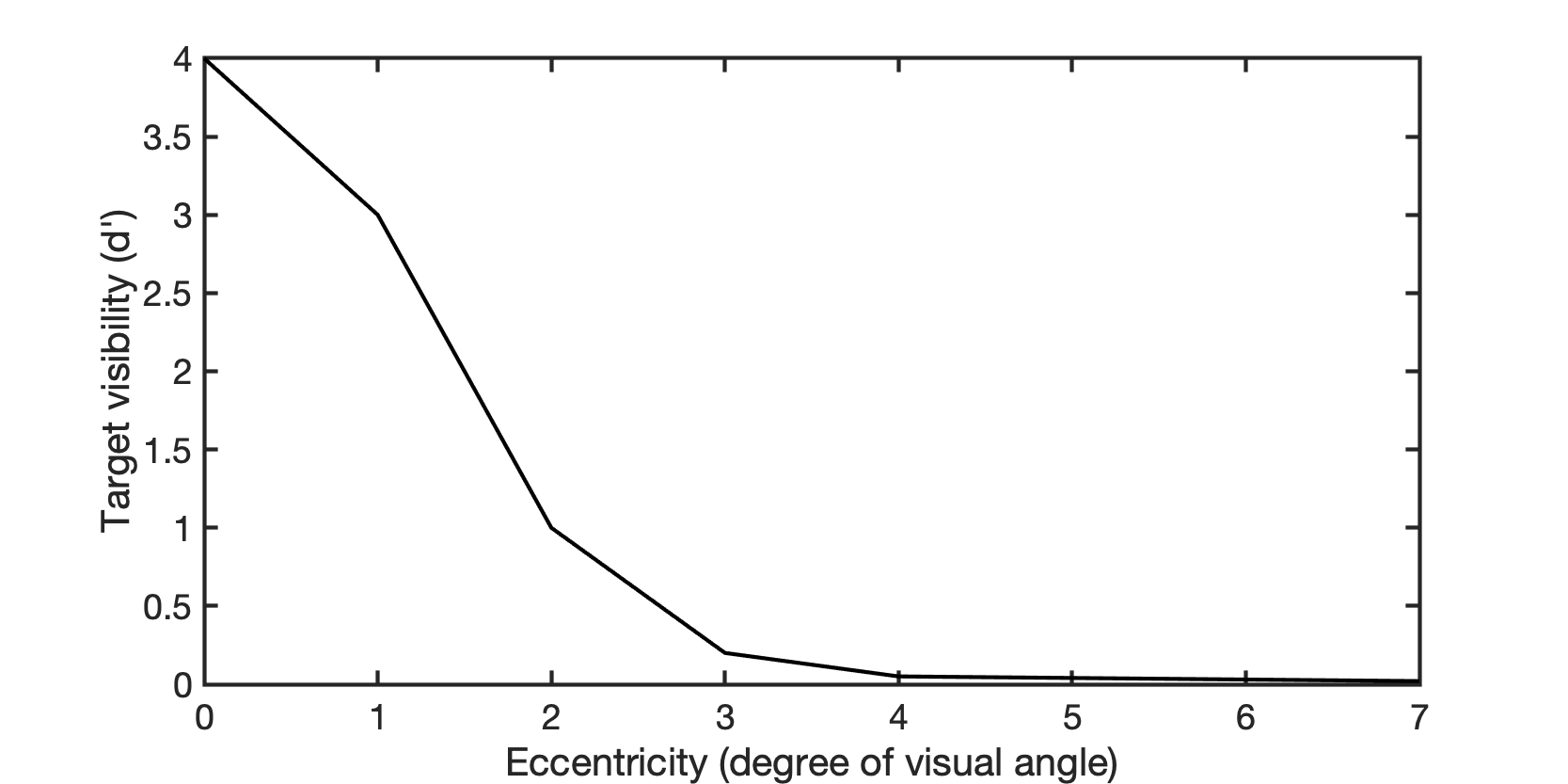}
         \vspace{0.2cm}
     \caption{The visibility of the target as a function of target distance to the fovea.}
      \label{fig:d}
   \end{figure}

We considered a visual search task that requires an observer to localize a signal that has $n=85$ possible locations covering a circular region in dynamic  (temporally uncorrelated) noise backgrounds. 
In this validation study, following Najemnik and Geisler, the template responses $W_{i, k(t)}$ were sampled from Gaussian distribution having the mean of 0.5 if the target is present at the $i^{th}$ location and, the mean of -0.5, otherwise. The standard deviation of $W_{i, k(t)}$ was  ${1}/{d'_{i, k(t)}}$, where $d'_{i, k(t)}$ denotes the visibility of the target at the $i^{th}$ location when the fixation is at the location $k(t)$. The value of  $d'_{i, k(t)}$ was determined according to the visibility map shown in Fig.~\ref{fig:d}.

In this preliminary study, the search models executed three eye movements that correspond to three new fixations after the initial fixation. The initial fixation was set to be at the center in the image. A fully-connected (FC) neural network having two FC layers was employed as the Q-network to learn Q values for each eye movement. Specifically, the first FC layer maps the state vector $\mathbf{s}$ having the dimension of 85 to a hidden vector having the dimension of 512 and the second FC layer maps the hidden vector to the vector of Q-value having the dimension of 85. A non-linear ReLU function is used in the first FC layer. 
When the Q-network (agent) was interacting with the environment to generate the training Q-value data, the template response at each one of the 85 locations was generated on-the-fly and the Q-value described in Eq. (\ref{eq:Q}) was empirically estimated.

\begin{figure}[H]
         \centering
         \includegraphics[width=0.8\textwidth]{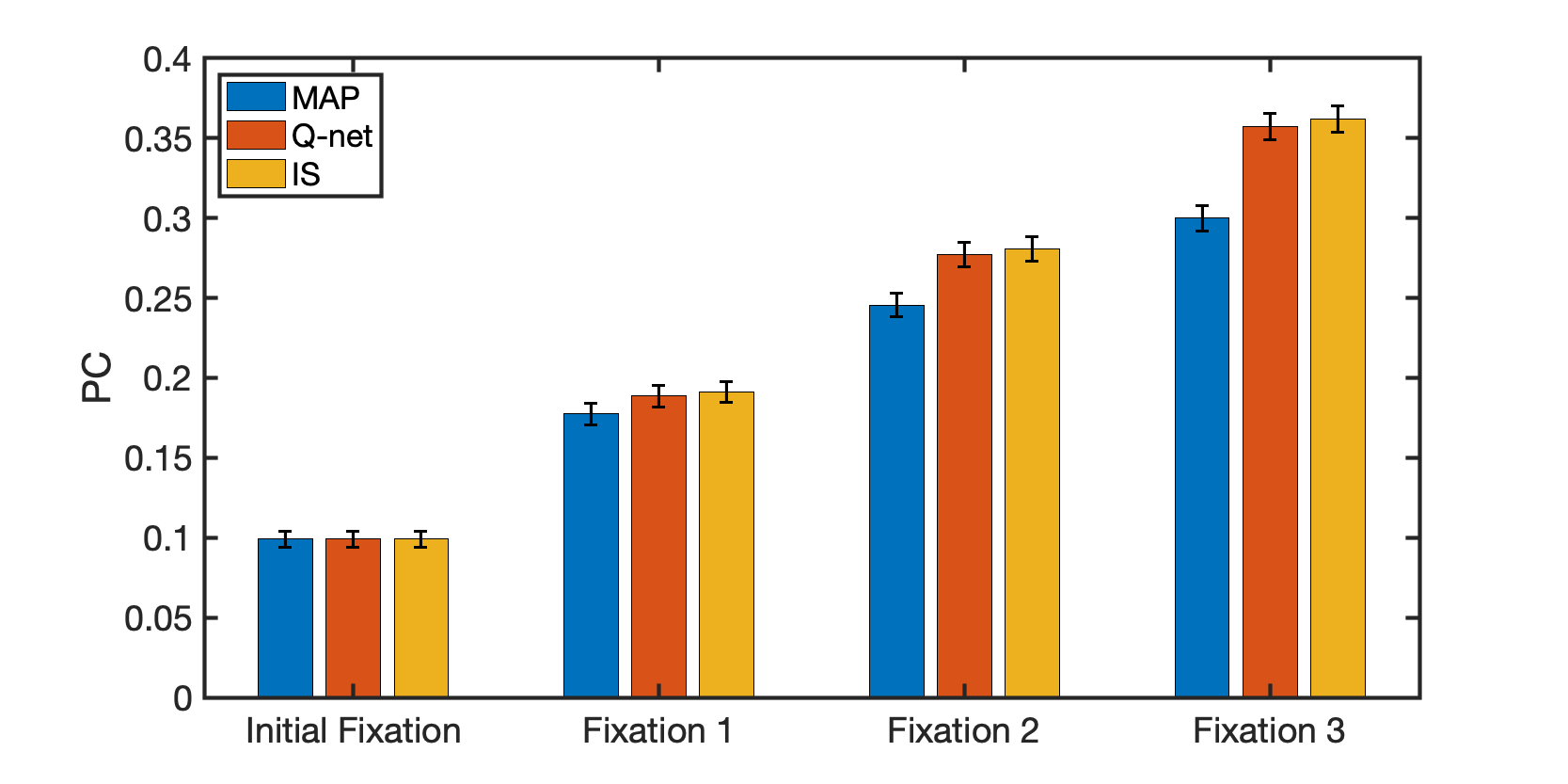}
     \caption{The proportion correct for the MAP searcher, ideal searcher, and Q-network searcher.}
      \label{fig:PC}
   \end{figure}

The search performance and spatial distribution of fixations corresponding to the proposed Q-network searcher were evaluated on a testing dataset having 3400 trials. 
The proportion correct of localization (PC) was employed as a metric to quantify the search performance.
The PC values produced by the MAP searcher, the ideal searcher, and the Q-network searcher are compared in Fig. \ref{fig:PC}.
The Q-network searcher achieved PC values close to the IS. Additionally, as expected, the Q-network searcher produced higher PC values than the MAP searcher.

\begin{figure} [ht!]
   \centering
 \hspace{-1cm}  \includegraphics[width=0.8\textwidth]{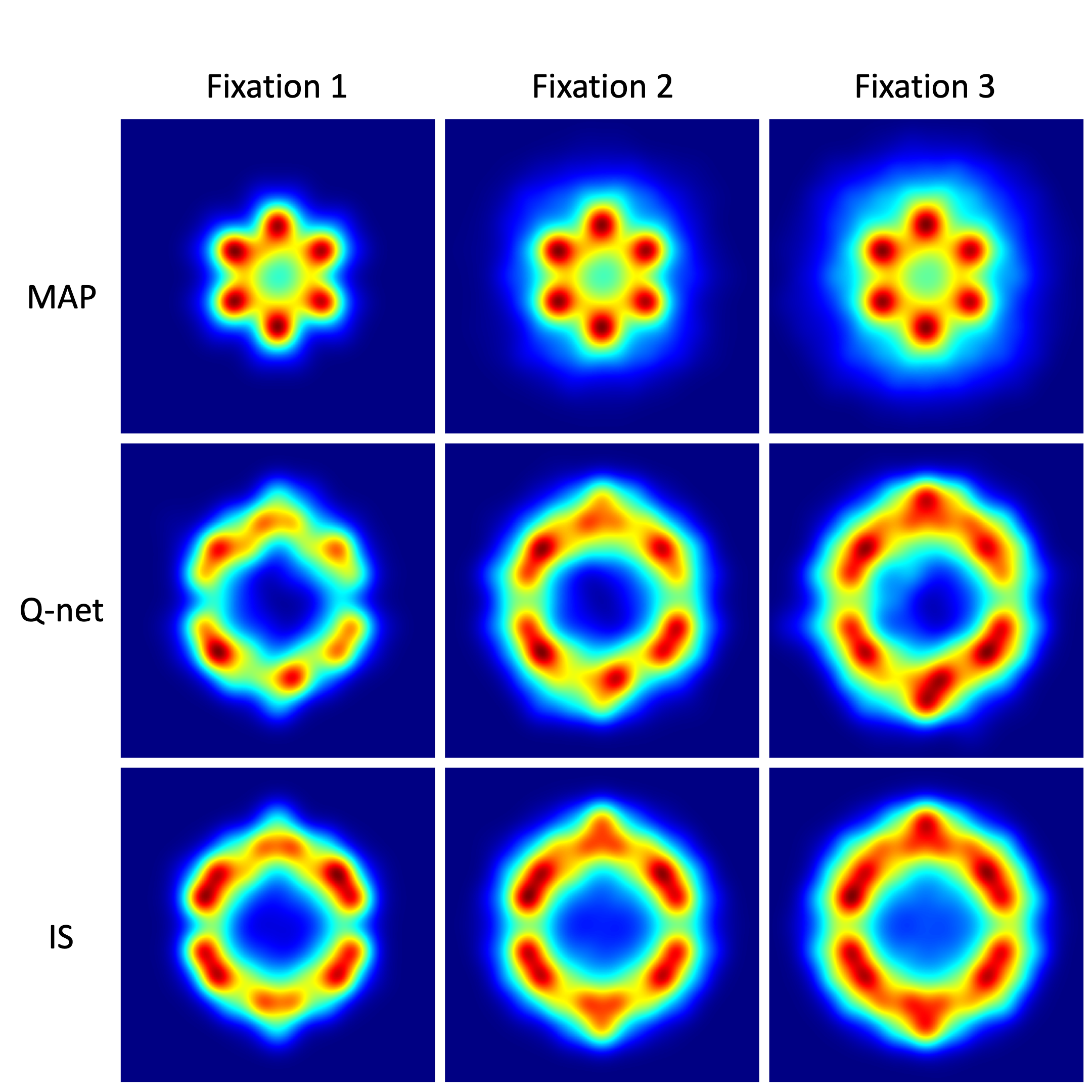}
 \vspace{0.2cm}
   \caption{
Spatial distributions of fixations corresponding to the MAP searcher (top row), Q-network searcher (middle row), and IS (bottom row).}
 \label{fig:fix} 
   \end{figure}

The spatial distributions of fixations produced by the three searchers are shown in Fig. \ref{fig:fix}.
The Q-network produced fixation distributions that are consistent with those produced by the IS, which are significantly different with the MAP searcher. 
   
\section{Discussion and Conclusion}
This work presents a reinforcement learning-based method that employs Q-network to approximate the Bayesian ideal searcher (IS) that optimizes the eye movement strategy for a foveated visual system. In this validation study, we considered a dynamic noise background case in which the template responses follow Gaussian distributions and the IS can be analytically determined. The proposed Q-network approach was implemented to execute three eye movements for the considered visual search task. The findings suggest that the Q-network searcher results in consistent eye movement behaviour and performance to the ideal searcher. To the best of our knowledge, this is the first study to investigate the use of reinforcement learning method for approximating the ideal searcher in vision.
Because reinforcement learning-based methods are data-driven and do not need to analytically determine the optimal decision strategy, our proposed Q-network approach is general and can be applied to cases in which the IS cannot be analytically determined.
Therefore, our method holds great potential to apply to clinically relevant vision tasks in which realistic anatomical backgrounds are considered. 
It will be important to validate the proposed method for more realistic tasks that requires a searcher to execute more eye movements in more complex backgrounds. Moreover, in our study, we employed separate Q-network for approximating the Q-value for each of the three saccades. This may become computationally inefficient when the search model is to be trained to execute a large number of eye movements. To address this potential limitation, one may employ a single large Q-network to approximate the Q-value for all saccades. The utilization of single large Q-network for approximating the IS that executes more saccades represents an important  future direction of research.  
\bibliography{report} 
\bibliographystyle{spiebib} 

\end{document}